\pdfoutput=1
\def\year{2022}\relax
\documentclass[letterpaper]{article} 
\usepackage{aaai22}  
\usepackage{times}  
\usepackage{helvet}  
\usepackage{float}
\usepackage{courier}  
\usepackage[hyphens]{url}  
\usepackage{graphicx} 
\usepackage{natbib}  
\usepackage{caption} 
\DeclareCaptionStyle{ruled}{labelfont=normalfont,labelsep=colon,strut=off} 
\usepackage[ruled,vlined]{algorithm2e}

\usepackage{newfloat}
\usepackage{listings}
\usepackage{multirow}
\usepackage{booktabs}
\usepackage[flushleft]{threeparttable}

\usepackage{amsfonts}
\usepackage{amsmath}
\usepackage{subcaption}
\usepackage{adjustbox}

\title{Learning to Solve Soft-Constrained Vehicle Routing Problems with Lagrangian Relaxation}

\author{
    Qiaoyue Tang\textsuperscript{\rm 1}\equalcontrib \footnote{Work done at Huawei Technologies Co., Ltd.},
    Yangzhe Kong\textsuperscript\equalcontrib \footnotemark[\value{footnote}],
    Lemeng Pan\textsuperscript{\rm 2},
    Choonmeng Lee\textsuperscript{\rm 2},
}
\affiliations{
    \textsuperscript{\rm 1}Department of Computer Science, The University of British Columbia\\
    \textsuperscript{\rm 2} Huawei Technologies Co., Ltd. \\
    qiaoyuet@cs.ubc.ca, kong.yangzhe@gmail.com, lee.choonmeng@huawei.com, panlemeng@huawei.com

}

\begin{document}

\maketitle
\begin{abstract}
Vehicle Routing Problems (VRPs) in real-world applications often come with various constraints, therefore bring additional computational challenges to exact solution methods or heuristic search approaches. The recent idea to learn heuristic move patterns from sample data has become increasingly promising to reduce solution developing costs. However, using learning-based approaches to address more types of constrained VRP remains a challenge. The difficulty lies in controlling for constraint violations while searching for optimal solutions. To overcome this challenge, we propose a Reinforcement Learning based method to solve soft-constrained VRPs by incorporating the Lagrangian relaxation technique and using constrained policy optimization. We apply the method on three common types of VRPs, the Travelling Salesman Problem with Time Windows (TSPTW), the Capacitated VRP (CVRP) and the Capacitated VRP with Time Windows (CVRPTW), to show the generalizability of the proposed method. After comparing to existing RL-based methods and open-source heuristic solvers, we demonstrate its competitive performance in finding solutions with a good balance in travel distance, constraint violations and inference speed.
\end{abstract}

\section{Introduction}
Vehicle Routing Problems (VRP) is a family of Combinatorial Optimization problems that finds the optimal routes for a fleet of vehicles to serve a set of customers. These problems have many industrial applications in the supply chain and logistic management scenarios, where efficiently designing the routes using appropriate optimization algorithms could significantly optimizing profits \cite{geir}. The unconstrained VRP generalizes the well-known Travelling Salesman Problem (TSP) with multiple vehicles, while the constrained VRPs extend the unconstrained ones with additional conditions. Among the different constraints, the hard constraints are strict in feasibility requirements, with constraint violation seen as having an infinite cost; on the other hand, the soft constraints often have a small tolerance level, i.e. a finite cost, on constraint violations \cite{soft_hard_constraint}. An examples of hard-constrained VRP is the VRP with pick-up and delivery (VRPPD) where packages are transported to specific customers or predetermined locations. Some other variants of the VRP can be seen as soft-constrained VRP, such as the capacitated VRP (CVRP) with packages transported using capacity-limited vehicles, and the VRP with time windows (VRPTW) with packages delivered following certain time schedules. The optimization goal in soft-constrained VRPs is to find the solution with the optimal target function subject to minimum cost.



Solving VRPs to optimality is known to be NP-hard, and the difficulty of the problem increases significantly with larger problem sizes and more complicated constraints \cite{toth}. Exact solution methods such as \textit{Branch-and-Cut} \cite{branch_and_cut} often suffer greatly from computational efficiency and are usually not applicable in real-world settings requiring instant planning. Heuristic methods are more effective but often require specific prior knowledge in exchange for good performances for each scenario. In recent years, with the fact that VRPs often share similar problem formulations and data structures, data-driven approaches that learn guided search patterns from successful experiences have become a popular alternative. In particular, Reinforcement Learning (RL) methods that require no explicit best-known solutions as labels and learn by self-supervising towards maximizing the cumulative gain become especially suitable in solving such NP-hard Combinatorial Optimization problems. 

Most existing work that use a learning-based approach in solving the routing problems has focused on the unconstrained TSP or VRP, or linearly constrained VRPs such as CVRP or SDVRP. Other variants of VRPs with a non-linear constraint or with multiple constraints are rarely discussed in the existing literature. The difficulty of such problems lies not only in the existence of non-linearity of constraints but also in performing the optimal solution search while simultaneously controlling for constraint violation costs. To fill the gap, we incorporate the Lagrangian relaxation technique \cite{numerical_optim}, a common approach in constrained optimization, into an RL-based framework in solving the soft-constrained VRPs. Specifically, the optimal solution is obtained by iteratively improving a fairly good solution towards the balanced optima that trades off the travel distance target and the constraint violation costs. Following the cumulative reward as a signal, the actions taken by the trained RL agent mimics the classic 2-exchange heuristic operations while being guided in the direction of optimal solutions. Our main contributions are:

\begin{itemize}
    \item We propose to solve the soft-constrained VRPs with an RL-based method that incorporates the Lagrangian relaxation technique for constraint violation control.
    \item We propose a generalized formulation for the soft-constrained VRPs with the Constrained Markov Decision Process (CMDP) and the appropriately designed optimization target and method to find closely optimal solutions.
    \item We apply the proposed framework on three types of soft-constrained VRPs, CVRP, TSPTW and CVRPTW. We demonstrate the effectiveness of the method by showing and analyzing its ability to find good-quality solutions with short inference time in multiple scenarios with varying problem sizes.
\end{itemize}




\section{Related Work}
\label{section:related_work}

\subsubsection{Learning-based Methods for VRPs}
Early learning-based work in solving non-constrained VRP (or mTSP) adopts the supervised learning approach. These work usually build a decision neural-network model and use known optimal solutions, often solved with exact methods, as supervising labels \cite{vinyals2017pointer, kaempfer2019learning}. The main drawback of supervised learning approaches is the difficulty in obtaining optimal solutions as labels, especially when the problem size is large upon which solution solvers become very slow or completely inapplicable \cite{toth}. 


Many recent work turn to the reinforcement learning approach where the supervision is given by well-designed reward signals. The solution strategy can be roughly grouped into (1) The end-to-end route construction approach where the model usually takes an encoder-decoder structure \cite{kool, xin2020, bello2017, nazari2018, Deudon2018, lin2021deep} or (2) Automating heuristic search by iteratively improving an existing solution with heuristic operators \cite{dai2018learning, rewriter, Lu2020, kwon2021pomo}. Most existing work use the policy optimization method where the model directly operates in the policy space to find optimal actions. Comparing to Q-value based methods such as DQN, policy gradients methods often have faster convergence rate, though controlling for variance when estimating the expected return is often needed for good training \cite{sutton_rl}. Variance reduction is often achieved by introducing an unbiased estimator of the target or an independent critic network. 

Many existing learning-based works target the TSP or CVRP problem. \citet{kool, xin2020} additionally solve the SDVRP and (Stochastic) Price Collecting TSP ((S)PCTSP) both of which have linear constraints on the decision variable that determines the solution (Appendix A). There have been a few work that attempted to solve time-window-constrained routing problems. \citet{Quentin} solve the TSPTW problem by integrating Dynamic Programming and RL which uses the results of it to reduce the action space of RL agents. \citet{zhang_twr} solves the TSPTWR (TSPTW with rejections) problem using an encoder-decoder based route construction approach and controls the constraint by rejecting constraint-infeasible customer nodes. \citet{lin2021deep} solves the EVRPTW (VRP with electric and time window constraints) using a encoder-decoder model structure and controls the constraint with a mix of masking and fixed penalty coefficients. \citet{zhang_multiagent} solves the CVRPTW problem in a multi-agent way which each agent constructs the route for one vehicle sequentially. It uses a similar encoder-decoder model as in \cite{kool} for each agent and similarly controls constraint violation by masking out infeasible nodes at each decoding step. 
Table \ref{related-work-vrps} summarizes a comparison of existing RL-based methods for solving VRPs.

\begin{table*}[h]
\centering
\begin{adjustbox}{max width=\textwidth}
\begin{tabular}{c|cccc}
\hline
 &
  \textbf{Approach} &
  \textbf{RL Training Method} &
  \textbf{Applied VRP Types} &
  \textbf{Constraint Handling} \\ \hline
\citet{dai2018learning} &
  \begin{tabular}[c]{@{}c@{}}Automate heuristic search\\ (insertion heuristic)\end{tabular} &
  \begin{tabular}[c]{@{}c@{}}Q-learning \\ (DQN)\end{tabular} &
  TSP &
  \textbackslash{} \\ \hline
\citet{bello2017} &
  Route construction &
  \begin{tabular}[c]{@{}c@{}}Policy Optimization \\ (Actor-Critic)\end{tabular} &
  TSP &
  \textbackslash{} \\ \hline
\citet{nazari2018} &
  Route construction &
  \begin{tabular}[c]{@{}c@{}}Policy Optimization\\  (Actor-Critic)\end{tabular} &
  CVRP, SDVRP &
  Masking \\ \hline
\citet{kool} &
  Route construction &
  \begin{tabular}[c]{@{}c@{}}Policy Optimization \\ (REINFORCE)\end{tabular} &
  \begin{tabular}[c]{@{}c@{}}TSP, CVRP, SDVRP, \\ PCTSP, SPCTSP\end{tabular} &
  Masking \\ \hline
\citet{rewriter} &
  \begin{tabular}[c]{@{}c@{}}Automate heuristic search \\ (2-exchange heuristic)\end{tabular} &
  \begin{tabular}[c]{@{}c@{}}Policy Optimization \\ (Actor-Critic)\end{tabular} &
  CVRP &
  \begin{tabular}[c]{@{}c@{}}Add vehicle in-place when \\ exceeding capacity\end{tabular} \\ \hline
\citet{Lu2020} &
  \begin{tabular}[c]{@{}c@{}}Automate heuristic search\\ (Large neighbourhood search)\end{tabular} &
  \begin{tabular}[c]{@{}c@{}}Policy Optimization \\ (REINFORCE)\end{tabular} &
  CVRP &
  \begin{tabular}[c]{@{}c@{}} Allow only feasible solution \\ in heuristic search\end{tabular} \\ \hline
\citet{xin2020} &
  Route construction &
  \begin{tabular}[c]{@{}c@{}}Policy Optimization \\ (REINFORCE)\end{tabular} &
  \begin{tabular}[c]{@{}c@{}}TSP, CVRP, SDVRP, \\ PCTSP, SPCTSP\end{tabular} &
  Masking \\ \hline
\citet{kwon2021pomo} &
  Route construction &
  \begin{tabular}[c]{@{}c@{}}Policy Optimization \\ (REINFORCE)\end{tabular} &
  \begin{tabular}[c]{@{}c@{}}TSP, CVRP \end{tabular} &
  Masking \\ \hline
\citet{lin2021deep} &
  Route construction &
  \begin{tabular}[c]{@{}c@{}}Policy Optimization \\ (REINFORCE)\end{tabular} &
  \begin{tabular}[c]{@{}c@{}} EVRPTW \end{tabular} &
  \begin{tabular}[c]{@{}c@{}}Masking + \\ fixed penalty coefficient\end{tabular}  \\ \hline
\citet{Quentin} &
  \begin{tabular}[c]{@{}c@{}}Route construction \\ (with Dynamic Programming) \end{tabular} &
  \begin{tabular}[c]{@{}c@{}}Q-learning (DQN); \\ Policy Optimization (PPO)\end{tabular} &
  TSPTW &
  \begin{tabular}[c]{@{}c@{}}Apply feasibility filter \\ on action space \end{tabular}  \\ \hline
\citet{zhang_multiagent} &
  \begin{tabular}[c]{@{}c@{}}Route construction \\ (one agent for one vehicle)\end{tabular} &
  \begin{tabular}[c]{@{}c@{}}Policy Optimization \\ (REINFORCE)\end{tabular} &
  CVRPTW &
  Masking  \\ \hline
\citet{zhang_twr} &
  Route construction &
  \begin{tabular}[c]{@{}c@{}}Policy Optimization \\ (REINFORCE)\end{tabular} &
  TSPTWR &
  \begin{tabular}[c]{@{}c@{}}Reject constraint-unsatisfied nodes + \\ Reward shaping \end{tabular}  \\ \hline
\textbf{Ours} &
  \begin{tabular}[c]{@{}c@{}}Automate heuristic search \\ (2-exchange heuristic)\end{tabular} &
  \begin{tabular}[c]{@{}c@{}}Constrained Policy Optimization \\ (REINFORCE)\end{tabular} &
  TSPTW, CVRP, CVRPTW &
  Lagrangian relaxation \\  \hline
\end{tabular}
\end{adjustbox}
\caption{Comparison of existing RL-based methods for solving Vehicle Routing Problems.}
\label{related-work-vrps}
\end{table*}




\subsubsection{Multi-objective Optimization in RL} 
Soft-constrained optimization problems can be transformed into multi-objective optimization problems with one main target and a series of constraints having different importance levels reflected by a set of penalty coefficients. Lagrangian relaxation methods which move the 'difficult' constraints into the target function is one of the common approaches for such problem \cite{numerical_optim}. In Reinforcement Learning theory, there are a lot of relevant work on Constrained Markov Decision Process (CMDP), and using Lagrangian multiplier is one of the main approaches to solve CMDPs \cite{BORKAR2005}. Using Lagrangian multipliers to balance the expected reward and expected cost of each constraint, \citet{BORKAR2005} proposes a two-timescale policy gradient method that extends the classic actor-critic method to the CMDP domain. Later, \citet{tessler2018reward} extends the work to a generalized form of constraints. Both work prove the theoretical property that the algorithm converges to finding a fixed-point feasible solution.

\section{The RL Model with Lagrangian Relaxation and Constrained Policy Optimization}
\label{section:model_whole}

\subsection{Problem Formulation}
\label{section:problem_form}
A Constrained Markov Decision Process (CMDP) can be defined by the tuple ($S, A, P, R, C$), where $S$ is the state space, $A$ is the action space, $P: S \times A \times S \mapsto [0, 1]$ is the state transition matrix, $R: S \times A \times S \mapsto \mathbb{R}$ and $C: S \times A \times S \mapsto \mathbb{R}$ are the immediate reward and cost functions when moving from one state to another under a given action. $C$ is the additional element in CMDP that represents (possibly a set of) cost functions.



The cumulative reward $G_{t}^{R} = f(R(s_{0}, a_{0}, s_{1}), ..., R(s_{t}, a_{t}, s_{t+1}), ...))$ and the cumulative cost $G_{t}^{C} = f(C(s_{0}, a_{0}, s_{1}), ..., C(s_{t}, a_{t}, s_{t+1}), ...))$ represent the collective value from a state onward, and can be defined as arbitrary functions of a series of immediate rewards and costs.


The optimization objective of CMDP is to find the optimal policy $\pi^{*}$, a probability distribution over actions, such that the expected cumulative reward is maximized while each expected cumulative cost is controlled within a threshold $\epsilon_{c}$, i.e.,

\begin{equation}\label{eq:cmdp}
\begin{gathered}
        \max_{\theta} \mathbb{E}_{\pi_{\theta}} \big[G_{R}(s)\big], \text{s.t. } \mathbb{E}_{\pi_{\theta}} \big[G_{C_{c}}(s)\big] \leq \epsilon_{c}, \forall c \in C,
\end{gathered}
\end{equation}

where $\pi_{\theta}$ is the policy $\pi$ parameterized by $\theta$, and $c$ is the index of constraints when there are multiple of them. 

Using the Lagrangian relaxation technique, the CMDP problem in Equation \eqref{eq:cmdp} can be converted into a regular MDP problem where the optimization goal becomes:

\begin{equation}\label{eq:mdp}
\begin{gathered}
    \min_{\lambda \geq 0} \max_{\theta} L(\lambda, \theta), \\
    L(\lambda, \theta) = \mathbb{E}_{\pi_{\theta}} \big[G_{R}(s) \big] - \sum_{c=0}^{|C|} \Big( \lambda_{c} \cdot (\mathbb{E}_{\pi_{\theta}} \big[G_{C_{c}}(s) \big] - \epsilon_{c}) \Big),
\end{gathered}
\end{equation}

where $L$ is the Lagrangian function, $\lambda \geq 0$ is a non-negative Lagrangian multiplier considering the inequalities in the cumulative cost function \cite{numerical_optim}. The problem in Equation \eqref{eq:cmdp} and \eqref{eq:mdp} has been shown to be equivalent \cite{altman_cmdp}.

\subsubsection{State, Action and Transition Probabilities}
We define \textit{State} to be a complete route sequence, e.g. $0 \to 1 \to 2 \to 3 \to 0 \to 4 \to 5 \to 0$ represents a route with two vehicles starting from the depot 0, serving customers 1, 2 and 3 (4 and 5 for the other vehicle) in order, and returning to the depot. An advantage to represent a solution as a sequence (e.g. over graph) is that the repeated depot node could be represented differently for each vehicle. The \textit{State Space} is all possible routes, i.e. given the number of customer $N_{c}$ and the number of vehicle $N_{v}$, $|S| = (N_{c} + N_{v} - 1)!/(N_{v} - 1)!$.  We train the agent to perform 2-exchange \textit{actions} (i.e. choose two nodes and swap their locations in the route) as in \citet{rewriter}. The \textit{action space} is all possible swaps that can be performed, $|A| = \binom{N_{c} + N_{v} - 1}{2}$. The underlying model is deterministic, i.e. given a state and action, the \textit{transition probability} to the next state is always 1, i.e. $P(s'|s, a) = 1, \forall (s, a, s')$. We note that both the state and action space is finite though extremely large especially with large $N_{c}$, therefore the optimal solutions surely exist but efficiently searching for the optima is challenging.

\subsubsection{Immediate Reward and Immediate Cost}
Let $\Delta_{s_{t} \to s_{t+1}}(f) = f_{t+1} - f_{t}$ represents the change of value when moving from state $s_{t}$ to $s_{t+1}$, $v_{\ell}$ represents the route length of vehicle $v$, $Dist(\cdot)$ calculates the travelling distance between two nodes. We define the immediate reward as the reduction of travelling distance, 



\begin{equation}\label{eq:reward}
    r_t = \Delta_{s_t \to s_{t+1}} \sum_{i=0}^{v_{\ell} - 1} Dist(i, i+1),
\end{equation}

Let $Cost(\cdot)$ represent the constraint specific violation cost, $N_{v}$ represent the number of vehicles and $|C|$ be the number of constraints. We define the immediate cost as the reduction of constraint violation cost, summed over all vehicles and all constraints if there are multiple of them, 


\begin{equation}\label{eq:cost}
    c_t = \Delta_{s_t \to s_{t+1}} \sum_{C} \sum_{v=0}^{N_{v}} Cost_{v, c}(s).
\end{equation}

\subsubsection{Constrained Cumulative Reward}

\begin{figure*}[h]
\centering
\includegraphics[width=1.7\columnwidth]{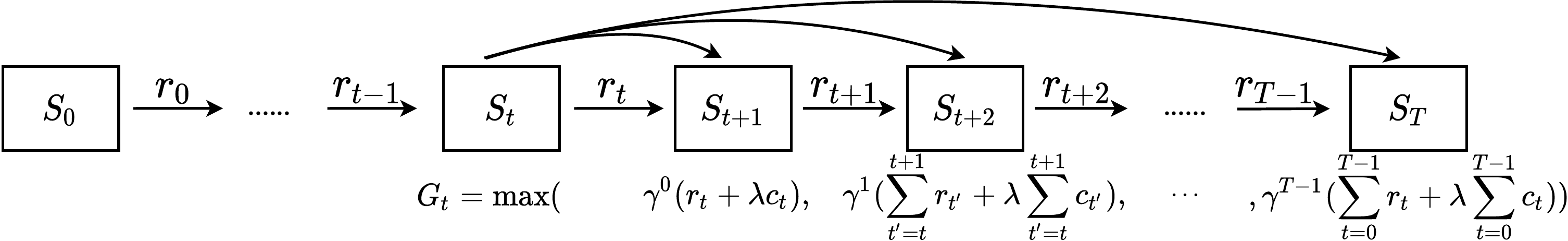}
\caption{Calculation of the Constrained Cumulative Reward $G_t$ take the maximization among all subsequent state pairs to improve convergence. Each term in the max operator can be derived as $\gamma^{t'-t-1}( \sum _{t=t}^{t'} r_t+\lambda \sum _{t=t}^{t'} c_t)=\Delta_{s_t \to s_{t'}} \sum_{i=0}^{v_{\ell} - 1} Dist(i, i+1)+\Delta_{s_t \to s_{t'}} \sum_{v=0}^{N_{v}} C_{v}(s)$. The decay coefficient discounts the steps between two states.}
\label{fig:G_s} 
\end{figure*}


By the linearity of expectation, the Lagrangian function in Equation \eqref{eq:mdp} can be written as, 

\begin{equation}\label{eq:mdp2}
\begin{gathered}
    L(\lambda, \theta) = \mathbb{E}_{\pi_{\theta}} \bigg[ G^R - \sum_{C} \Big( \lambda_{c} \cdot (G^{C_{c}} -  \epsilon_{c}) \Big) \bigg], 
\end{gathered}
\end{equation}

where the expression inside the expectation can be seen as an arbitrary function balancing the cumulative reward and cost. Without explicitly defining $G_{t}^{R}$ and each $G_{t}^{C}$, We define the \textit{Constrained Cumulative Reward} ($G_{t}$) directly. 

Let $S_T$ represent the terminal state of a sample trajectory $(s_{0}, a_{0}, r_{0}, c_{0}; \ldots; s_{T}, a_{T}, r_{T}, c_{T})$, and let $\gamma$ represent the decay coefficient, 



\begin{equation}\label{eq:cum_constrained_reward}
\begin{gathered}
    G_t = \max_{t'} \bigg( \gamma^{t'-t} \Big( \sum_{i=t}^{t'} \big(r_i -  \sum_{C} \lambda_{c} \cdot (c_i - \epsilon_{c}) \big) \Big) \bigg), \\
    t = 0, \ldots, T.
\end{gathered}
\end{equation}

Then the optimization goal can be rewrote as, 

\begin{equation}\label{eq:optimize_goal}
    \min_{\lambda \geq 0} \max_{\theta} \mathbb{E}_{\pi_{\theta}} [G_t].
\end{equation}

We provide an intuition for designing $G_{t}$. $G_t$ plays the same role as the \textit{Return} function (an exponentially discounted sum of immediate rewards) in classic RL problems, in which they both are the signal to guide the agent's selection of actions. The expression of $G_{t}$ is specially designed to encourage better performance in soft-constrained VRPs with 2-exchange moves. First, the immediate reward is penalized by the immediate cost such that the agent is encouraged to find better moves while balancing the reward and cost with iteratively updated $\lambda$s. In addition, We calculate the cumulative value using the maximum value of all pairs of subsequent moves from $s_{t}$ to $s_{t'}$ instead of a summation over all consecutive moves from $s_{t}$ to $s_{t+1}$ as in the $Return$ definition. "Bad" operations that do not improve the objective function will be suppressed, while only the 'good' actions are rewarded with the $\max$ function. It also tends to decorrelate the effect of a series of historical operations so that the agent is less affected by locally optimal trajectories. Furthermore, the decay coefficient $\gamma$ discounts more for more distant state pairs because the further state becomes more independent (less relevant) of the past as more 2-exchange operations are performed along the way. To sum up, we apply such modification to better mimic the heuristic search process by encouraging more immediate and effective actions that improve the cost-penalized objective function. Figure \ref{fig:G_s} provides a visual representation of the definition of $G_t$.


\subsection{Model}
\label{section:model}
Though the majority of existing frameworks use the end-to-end route construction approach, it might be difficult to apply to our target problems for the following reasons: (1) It is difficult to design an efficient masking scheme to consider both the fundamental constraints in plain VRP (e.g. each node can only be visited once) and the additional constrains (e.g. capacity, time window). A complicated design of the masking scheme is needed to avoid violations of multiple constraints simultaneously as shown in \citet{lin2021deep}. Checking constraint violations itself could also be computationally expensive. For example, let $n$ and $\ell$ represent the number of customers and route length, the complexity of obtaining the set of feasible nodes at all decoding steps are $O(n^{2}\ell)$ for both the capacity and time window constraints. (2) Applying masking scheme is equivalent to applying a strict feasibility filter on the action space. When feasibility checks are applied to sequence decoding procedures, sequentially constructed solutions may easily fall into local optima due to the limited exploration space of agents. (3) When training end-to-end route construction models with policy gradient algorithms, the model only receives the reward signal after constructing a complete solution, i.e. the reward signal contains an aggregated signal for constructing the whole route instead of an immediate signal for each decoding step. Therefore, end-to-end route construction models work better with masking instead of penalizing in the form of Lagrangian relaxation due to the lack of guidance at single decoding steps. 

\begin{figure*}[ht]
\centering
\includegraphics[width=1\textwidth]{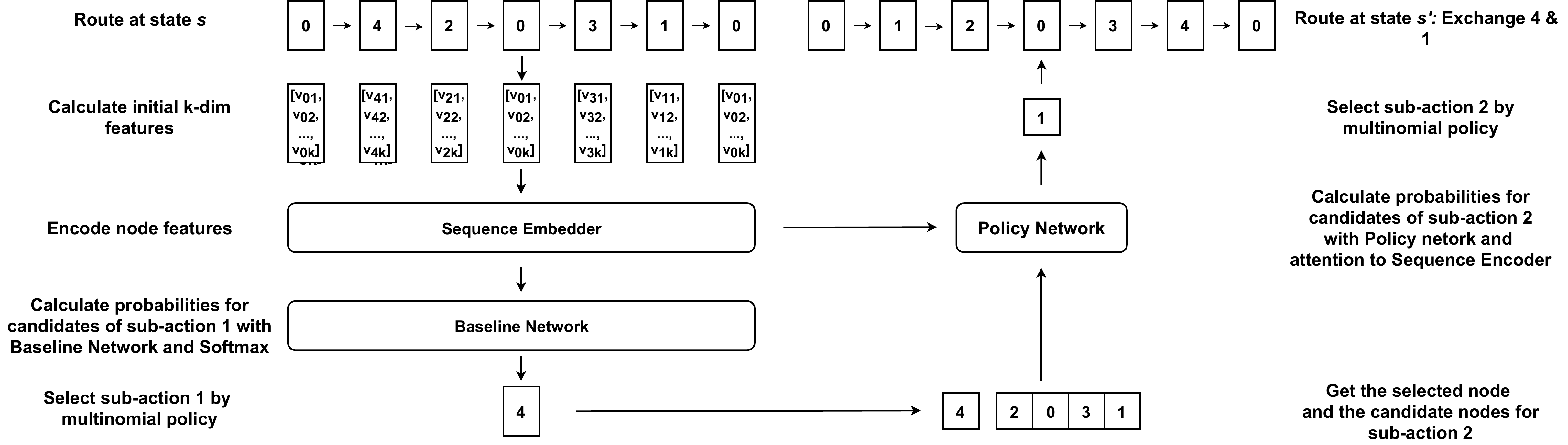}
\caption{The model structure. Given State $S{t}$, the model takes the following steps to choose an action. (1) Baseline network calculates the Q-value of candidates sub-action 1. (2) The model chooses the first node (sub-action 1) according to the Q-values. (3) Obtain the action space of sub-action 2 given the first node. (4) Policy network chooses the second node (sub-action 2). (5) Exchange the first node and the second node to get the next state $S_{t+1}$. We note that the Baseline Network used in sub-action 1 is also reused as the baseline in REINFORCE algorithm to calculate the policy loss for Policy Network.}
\label{fig:agent}
\end{figure*}

Taking the above into consideration, we apply Lagrangian relaxation to the automating heuristic search approach. We refer to and adopt the model architecture of \citet{rewriter}. Our model iteratively updates an existing solution by training an RL agent to perform the 2-exchange action. The complete action of exchanging two nodes is split into two sub-actions performed sequentially. The agent selects the first node based on the estimates of the baseline network. The second node is selected based on a policy network that takes in the integrated information of the first node and its possible swapping candidates, calculated through a dot-product attention mechanism.
Figure \ref{fig:agent} shows the model architecture we use for training.

\subsubsection{Trajectory Shaping}

We further improve the model performance by intervening the trajectory generation process to boost the quality of the agent's training information. The motivation is similar to modifying the expression of $G_{t}$. Due to the large search space and the sparsity of optima, guiding the agent to explore and learn the 'good' actions can be very slow or easily trapped into local optima, especially if the initial state solution is far from the true global optimum. With the underlying model being deterministic and we can easily obtain the next state's reward and cost, we suggest a post-action rejection rule to modify the generated trajectories. 

After the policy network selecting an action $a_{t}$ at state $s_{t}$ and transitioning to the next state $s_{t+1}$, we check whether the solution is improved, i.e. $\sum Dist(i, i+1|s_{t}) + Cost(s_t) \geq \sum Dist(i, i+1|s_{t+1}) + Cost(s_{t+1})$. we reject adding ($s_{t+1}, a_{t+1}, r_{t+1}, c_{t+1}$) to the trajectory with probability $\phi$ and $1-\phi$ respectively when non-improved and improved solutions are found,


\begin{equation}
    P(\textnormal{Reject}) = \left\{
        \begin{array}{ll}
            \phi & \quad \textnormal{if improved} \\
            1 - \phi & \quad \textnormal{if not improved},
        \end{array}
    \right.
\end{equation}

where empirically $\phi \in [0, 1]$ can start from a larger value and decay gradually to $\epsilon$ during training, and stay fixed afterwards. When a rejection takes place, we return to the state $s_t$ and sample a new tuple of ($s_{t+1}, a_{t+1}, r_{t+1}, c_{t+1}$) until the new experience is accepted into the trajectory.

We note that modifying the experience learned by agents is quite common in RL literature to deal with problems such as sparse reward \cite{her}. In our case, applying a rejection rule can be seen as an efficient way to reduce the search space where it is less likely to find optimal solutions. It speeds up the agent's training process, especially at early stages, by showing agents good sample trajectories that possibly lead to better solutions. Instead of having a deterministic rejection rule, the exploration mechanism in RL is kept by assigning a positive value $\phi$ as the rejection probability. We make $\phi$ gradually reduce and stabilize to $\epsilon$ so that there is less intervention at late training stages. $\epsilon$ is a tunable hyperparameter and has a similar role as in classic $\epsilon$-greedy action selection.

\subsection{Training}
We train the model following the \textit{Constrained Policy Optimization} approach \cite{tessler2018reward}, applying multiple constraints and a baseline to our context. The goal described in Equation \eqref{eq:cum_constrained_reward} and \eqref{eq:optimize_goal} can be optimized with a multi-timescale approach where the Lagrangian coefficients $\lambda$s are considered fixed while maximizing over $\theta$, then $\lambda$s are optimized next with the current policy $\pi_{\theta}$. The inner step of maximizing with respect to policy network parameter $\theta$ can be optimized iteratively using standard gradient ascent. To stabilize training, we use a parameterized network, $V_{\omega}(s)$, that takes in \textit{state} and estimates $\mathbb{E}_{\pi}[G_{t}]$, as the baseline as in the \textit{REINFORCE} algorithm \cite{reinforce}. 



The baseline network essentially performs value function approximation in a Monte-Carlo way. We note that the baseline is considered as a \textit{Critic} because no bootstrapping is used to update its value. The parameters of the baseline network are updated as follows, 

\begin{equation}\label{eq:baseline_update}
\begin{gathered}
    \delta = G_{t} - \hat{v}_{\omega}(s_{t}), \\
    \omega \leftarrow \omega + \alpha_{\omega} \delta \nabla_{w} \hat{v}_{\omega}(s_{t}),
\end{gathered}
\end{equation}

where we calculate the Mean Squared Error between the estimated value from the baseline network and the actual $G_t$ as the loss (MSE Loss),

\begin{equation}
\begin{gathered}
    MSE(\omega) = \mathbb{E}_{\omega} \big[ \delta^{2} \big].
\end{gathered}
\end{equation}

Following the \textit{Policy Gradient Theorem} \cite{sutton_rl}, the policy network parameters are updated as, 

\begin{equation}\label{eq:theta_update}
\begin{gathered}
    \theta \leftarrow \theta + \alpha_{\theta} \nabla_{\theta} \mathbb{E}_{\pi_{\theta}}[\delta], \\
    \nabla_{\theta} \mathbb{E}_{\pi_{\theta}}[\delta] = \mathbb{E}_{\pi_{\theta}} \Bigg[G_{t} \sum_{t=0}^{T} \nabla_{\theta} \log \pi_{\theta}(a_{t}|s_{t}) \Bigg].
\end{gathered}
\end{equation}

$\lambda$ can be optimized using subgradient descent, since the maximizer $\Tilde{\theta} \in \arg \max_{\theta} \mathbb{E}[G_{t}] $ is non-unique for an arbitrarily fixed $\lambda$, the Lagrangian function is non-differentiale at $\Tilde{\theta}$ \cite{numerical_optim}. With the assumption that the time is discrete in each trajectory, i.e. $t \in \mathbb{Z}: 0 \leq i \leq T$, the Lagrangian multipliers are updated as, 

\begin{equation}\label{eq:lambda_update}
\begin{gathered}
     \lambda \leftarrow \max \Big(0, \lambda - \alpha_{\lambda} \nabla_{\lambda} \mathbb{E}_{\pi_{\theta}}[\delta] \Big), \\
    \nabla_{\lambda_{c}}\mathbb{E}_{\pi_{\theta}}[\delta] = - \mathbb{E}_{\pi_{\theta}} \bigg[\gamma^{\tilde{t}-t} \sum_{i=t}^{\tilde{t}} (c_{i} - \epsilon) \bigg], \\
    \tilde{t} = \mathrm{argmax}_{t'} \bigg( \gamma^{t'-t} \Big( \sum_{i=t}^{t'} \big(r_i -  \sum_{C} \lambda_{c} \cdot (c_i - \epsilon) \big) \Big) \bigg).
\end{gathered}
\end{equation}


The pseudo code for one update of the training process is illustrated in Appendix A.



\section{Experiments}
\label{section:experiments}

We empirically demonstrate the effectiveness of the proposed method by applying it to three types of soft-constrained VRPs: TSPTW, CVRP, and CVRPTW. We use simulated datasets to train and evaluate the model performances. We compare our method with both open-source heuristic solvers and other RL-based approaches. We also experiment on the initial value of Lagrangian coefficient $\lambda$ and the effectiveness of adopting the designed return function and trajectory shaping.

Following the general form described in Section \ref{section:problem_form}, we define the constraint violation costs for the time window and capacity constraints respectively. Let $Time(\cdot)$ calculate the travelling time between two nodes, and let $[tw_{i, 0}, tw_{i, 1}]$ be the $[\textnormal{start, end}]$ value of time windows of node $i$, the time-window cost is defined as, 

\begin{equation}\label{eq:vrptw_cost}
\begin{split}
    Cost_{v, tw} &= \textnormal{Early arrival cost} + \textnormal{Late arrival cost} \\ 
    &= \sum_{j=1}^{v_{l}-1} \Bigg[ \max \Big(\sum_{i=0}^{j-1} Time(i, i+1) - tw_{j, 0}, 0 \Big) \Bigg]\\
    &+ \sum_{j=1}^{v_{l}-1} \Bigg[ \max \Big(tw_{j, 1} - \sum_{i=0}^{j-1} Time(i, i+1), 0 \Big) \Bigg].
\end{split}
\end{equation}

Let $Dmd_{i}$ represent the demand of node $i$ and let $Cap_{v}$ be the capacity of vehicle $v$, the capacity cost is defined as,

\begin{equation}\label{eq:cvrp_cost}
    Cost_{v, cap} = \frac{\sum_{i=1}^{v_{\ell}-1} Dmd_{i} - Cap_{v}}{Cap_{v}}.
\end{equation}

We normalize the excessive load to the same scale as the distance travelled by dividing it by capacity.

\begin{table*}[ht!]
\large
\centering
\begin{adjustbox}{max width=\textwidth}
\renewcommand{\arraystretch}{1.10}
\begin{tabular}{cc|cccc|cccc|cccc}
 & & \multicolumn{4}{c|}{N = 20} & \multicolumn{4}{c|}{N = 50} & \multicolumn{4}{c}{N = 100} \\
 & Method & Obj. & Tgt. & Cost & Time(s) & Obj. & Tgt. & Cost & Time(s) & Obj. & Tgt. & Cost & Time(s) \\ \hline \hline
\multirow{8}{*}{\rotatebox[origin=c]{90}{CVRP}}& Random    & 13.65(1.29)  & 12.96      & 0.69    & 0.03    &  31.43(2.25)    &29.68      &    1.75  &     0.06    &  60.96(3.62)    &  58.16    &   2.81   &    0.09     \\\cline{2-14}
                        & LKH3                 & 6.12(0.79)  & 6.12      & 0.00      & 9.78    & 10.35(1.23)     &   10.35   &  0.00    &    33.31     &  15.66    & 15.66     &   0.00   &    61.63     \\
                        & OR Tools             & 6.58(0.72)  & 6.58      & 0.00      & 0.16    &  11.28(1.10)    &   11.28   & 0.00     &    0.58     &  17.20    &   17.20   &   0.00   &   2.38      \\\cline{2-14}
                        & NR                   & 6.16      & 6.16      & 0.00     &    &    10.51     &     10.51 &   0.00   &      &     16.10    &    16.10  &     0.00 &              \\
                        & AM                   & 6.25      & 6.25      & 0.00      & 120     &  10.62 & 10.62   &   0.00   &   1680   &    16.23     &   16.23 &  0  &   7200          \\
                        & RL                   & 6.40      & 6.40      & 0.00      &     &  11.15  & 11.15     & 0     &      &  16.96       &  16.96    &   0   &               \\
                        & L2I                  & 6.12      & 6.12      & 0.00      & 720     &   10.35   &  10.35    &  0    &      1020   &    15.57  &    15.57  &    0  &    1440     \\
                        & \textbf{Ours}        &   6.32(0.75)        &    6.29       &    0.03       &    42.62     &   11.61(1.06)   &  11.48    &  0.13    &   67.28     &   17.17(1.00)   &  16.80    &  0.37    &     116.81    \\ \hline\hline
\multirow{4}{*}{\rotatebox[origin=c]{90}{TSPTW}} & Random  & 75.94(12.99) & 10.70+0.00   & 65.24     & 0.04    &   525.19(53.20)   &    26.77+0.00  &   498.42   &    0.12     &  2219.95(147.40)    &  52.87    &  2167.08    &   0.45      \\\cline{2-14}
                        & LKH3                 & 10.32(1.27)     & 6.91+3.41 & 0.00      & 16.27   &  25.02(1.94)    &  12.46 + 12.56    &    0.00  &     64.35    &    51.09(2.78)  &  22.22+28.87    &    0.00  &     194.46    \\
                        & OR Tools             & 10.59(1.31)     & 6.92+3.67 & 0.00      & 32712.72    &  NA    &  NA    &    NA  &  $\textnormal{NA}^{*}$       &  NA    &  NA    &  NA    &  $\textnormal{NA}^{*}$   \\\cline{2-14}
                        & \textbf{Ours}        &     10.05(1.29)      &   10.05        &  $<$ 0.01         &     36.58   &    24.39(1.95)  &  24.39    &  $<$ 0.01    &    66.85     &  52.52(2.65)    &  51.69    &  0.83    &    117.58     \\ \hline\hline
\multirow{5}{*}{\rotatebox[origin=c]{90}{CVRPTW}} & Random & 35.66 (13.60)  & 12.96+0.00    & 0.69+22.01 & 0.08    &    140.67(42.37)  &    29.68+0.00  & 1.75+109.24     &   0.20      &   372.81(102.42)   &   58.16   &  2.81 +311.85    &    0.47     \\\cline{2-14}
                        & LKH3                 & 6.29 (0.60)   & 6.21+0.08 & 0.00+0.00 & 20.91   &  11.39(1.29)    &  11.15+0.24    &   0.00+0.00   &   95.62      & 16.70(2.20)     &   16.15+0.55   &   0.00+0.00   &   238.94      \\
                        & OR Tools             & 7.15(0.71)      & 6.90+0.25 & 0.00+0.00 &   1.27      &  12.27(0.98)    &    11.76+0.51  &  0.00+0.00   &  2.31 &    19.81(1.40)     & 18.23+0.95     &   0.00+0.00   &   5.08           \\\cline{2-14}
                        & \textbf{Ours (TW)}   &  6.46(0.66)         &    6.45       &    $<$0.01       &    63.09     &    12.15(1.07)  & 12.06+0.00     &    0.00+0.09  & 77.16 &   19.79(2.08)    &     19.41 + 0.00 & 0.00+0.38     &       129.86     \\
                        & \textbf{Ours (Both)} &     6.64(0.62)      &    6.60+0.00       &   0.01+0.03        &   66.84      &    13.29 (0.53) &12.92+0.00      &   0.02+0.35   &    79.32     &     21.69(1.62) &     19.65+0.00 &    0.03+2.01 &     134.37    \\
\end{tabular}
\end{adjustbox}
\caption{The result table comparing the performance of our model with other baselines. The objective (Obj.) is a sum of the target (Tgt.) and Cost. Values in the Obj. column is the mean (sd) over problems in the test set. (1) CVRP: Each vehicle's capacity is $30, 40, 50$ in $N=20, 50, 100$ cases respectively. Values in Tgt. and Cost column are total travel distance and capacity violation cost. (2) TSPTW: Values in Tgt. column are total travel time + waiting time, values in Cost column are time window violation cost. (3) CVRPTW: Values in Tgt. column are total travel time + waiting time, values in Cost column are capacity + time window violation cost. *NA in TWPTW for $N=50, 100$ because it ran $>24$ hours for a single instance.}
\label{table:results_baseline}
\end{table*}

\subsection{Results}

\subsubsection{Baselines}
We compare the results run by a trained agent versus an agent following the random policy (\textbf{Random}), i.e. randomly exchanging 2 nodes to obtain new solutions. We also compare our results with two open-source heuristic solvers, \textbf{OR Tools} \cite{ortools} and \textbf{LKH3} \cite{lkh3}. The OR Tools is set to run once with 60 seconds time limits for each number of vehicles. LKH is run once with the maximum number of trails set to 10000. For the widely studied CVRP problem, we additionally compare our method with the Neural Rewriter model (\textbf{NR}) \cite{rewriter}, the attention model (\textbf{AM}) \cite{kool}, the pointer-network model (\textbf{RL}) \cite{nazari2018}, and the Learning to Improve model (\textbf{L2I}) \cite{Lu2020}. We do not retrain these baseline, and the test set results are either run with author-provided checkpoint or directly using the author reported results. The time-window-constrained VRPs are less investigated in which the problem types vary and data are often not publically available. Therefore, we mainly compare our results with heuristic methods and only make qualitative comparisons with other existing learning-based work. More training and inference details are presented in Appendix A. 

\subsubsection{Metrics}
We evaluate the performance using the following metrics: (1) \textbf{Obj.}: We calculate the objective value as the sum of the target and costs. With both lower target and lower costs being favored, we formulate the objective function such that they have the same weight of 1 to provide a direct comparison of these methods. However, the user may define different objective functions with different weights to evaluate their preference over balancing the target and costs. (2) \textbf{Tgt.}: The value of the target function being optimized. The target function is the summed travel distance of all vehicles in CVRP and is the summed travel time of all vehicles in TSPTW and CVRPTW. (3) \textbf{Cost}: The summed constraint violation cost of all vehicles as calculated in Equation \ref{eq:vrptw_cost} and \ref{eq:cvrp_cost}. (4) \textbf{Time(s)}: The inference time to solve a single problem instance. For a clear decompose of \textbf{Tgt.} and \textbf{Cost}, we separately report the total travelling distance/time, the cost of exceeding capacity, the early arrival and late arrival cost. In all cases, we report the mean score (standard deviation for the objective value additionally) for all metrics across 100 randomly selected problem instances. A lower value in all metrics indicates better performance.

\subsubsection{Discussion}
Table \ref{table:results_baseline} presents the results of our model and the baselines. For all cases, we obtain a lower objective value than \textit{Random}. It indicates that the learned policy is at least informative to find better solutions. Our approach typically finds competitive solutions with small target values while maintaining small constraint violation costs compared to baselines. Such advantage is especially strong in time-window constrained problems. The solution quality remains almost the same as problem size increases, with only a slow increase in the inference time.

In CVRP problems, our results are roughly at the same level as the other methods. The baseline methods all use hard-coding for the capacity constraint so the costs are zero. We treat it as a soft-constraint problem but the cost is well-controlled with all values $<1$. We have the advantage over several RL-based approaches in terms of inference time especially comparing to AM and L2I. A potential room of improvement for the sparse cost in CVRP could be exchanging nodes between routes of different vehicles only and optimizing the target function within a single vehicle using TSP methods in an iterative way.

In the TSPTW problem, the optimization goal in both LKH3 and OR Tools is the overall travel time including the waiting time.For a fair comparison, we consider waiting time as a part of the optimization objection and not as a cost. \footnote{Both heuristic solvers have fairly strict format requirements, which leaves the user with very limited space to modify the objective function. The waiting time option can be set to 0 in OR Tools but still it outputs solutions with nonzero waiting time.
} We have strong advantage for the $N=20, 50$ datasets where we find solutions with a lower target value while having almost negligible cost. In the $N=100$ dataset, we still find solutions with smaller target value and shorter inference time comparing to LKH3. The cost could potentially be reduced more if we allow both longer training and inference time. We qualitatively compare our result with \cite{Quentin, zhang_twr} since there are some differences in problem type and result reporting strategy. \citet{Quentin} reports their results with the binary status of success/fail and optimal/non-optimal. Our model may have an advantage in inference time as they reported time-out in some cases. \citet{zhang_twr} allows rejecting service to infeasible customer nodes which is a different problem setting and could lead to shorter travel time and lower costs.

We experiment on two implementations for the multi-constraint CVRPTW problem. In the (TW) implementation, we only apply Lagrangian relaxation to more difficult time window constraint, while the capacity constraint is hard-coded by adding a vehicle in place once capacity is exceeded. In the (Both) implementation, both constraints are Lagrangian-relaxed using two different sets of Lagrangian multipliers. Comparing the (TW) results to the baselines, we find good quality solutions with similar target values, small costs with all values $<0.5$ and shorter inference time in larger problem size datasets ($N=50, 100$). We qualitatively compare our results with two closely related works of \cite{zhang_multiagent, lin2021deep}. Our results show that the gap between the solutions found by the model and LKH3 ($3\%, 6\%, 18\%$ for $N=20, 50, 100$) is smaller than the results reported in these two works (between $8\% \sim 20\%$). Though hard-coding the capacity constraint gives better results, we show that it is feasible to optimize two constraints simultaneously as both the target and the costs are much smaller than \textit{Random}. Possible ways to improve results may include longer training time and increasing the number of exchange steps since CVRPTW is inherently more complex.

\begin{figure}[hb!]
     \centering
     \begin{subfigure}[b]{0.4\textwidth}
         \centering
         \includegraphics[width=\columnwidth]{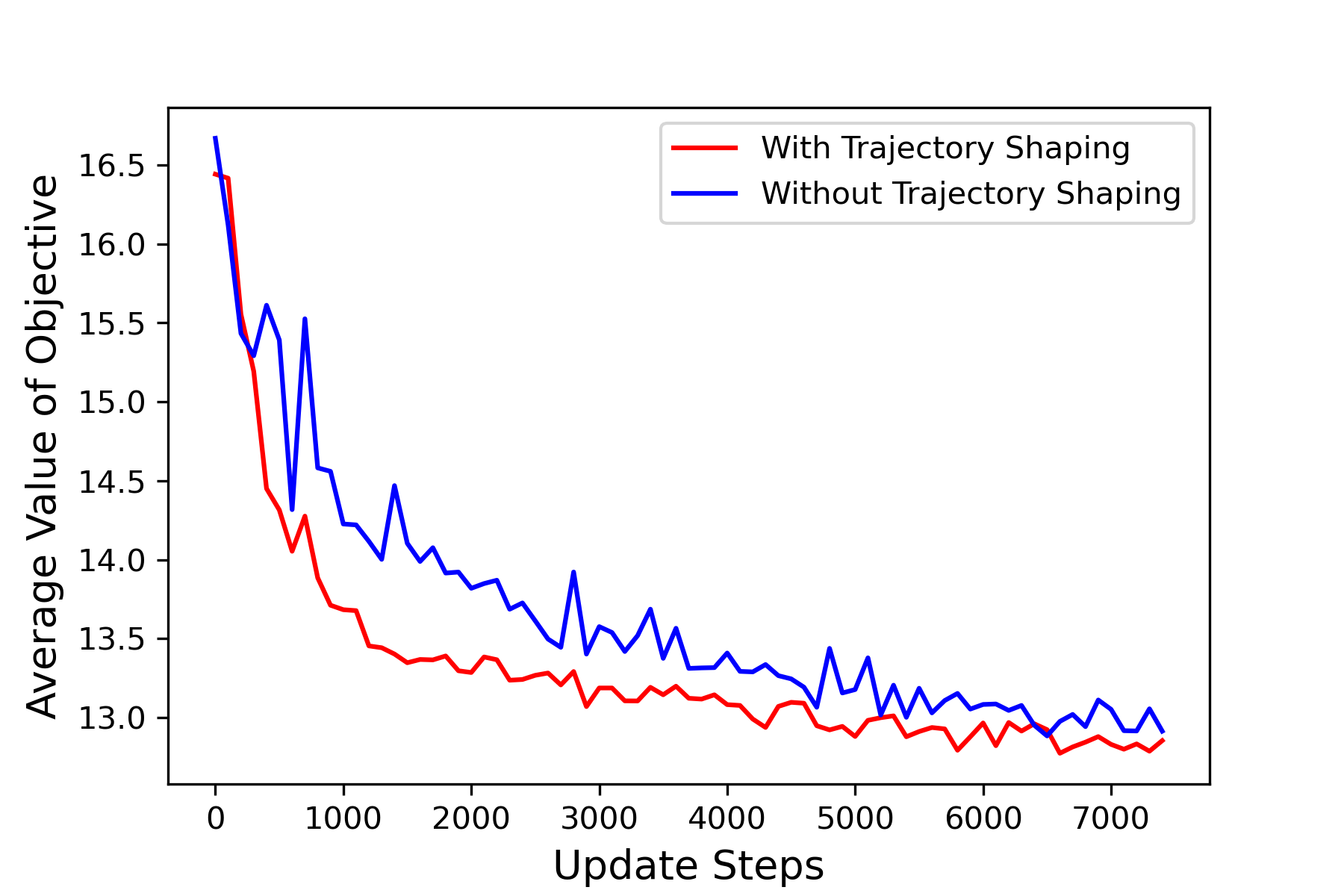}
         \caption{First 4000 update steps of our model with and without Trajectory Shaping on the dataset of CVRPTW-50. Although training without Trajectory Shaping could lead to similar results, training with Trajectory Shaping shows a higher convergence rate.}
         \label{fig:tc_vs_no_tc}
     \end{subfigure}
     \hfill
     \begin{subfigure}[b]{0.4\textwidth}
         \centering
         \includegraphics[width=\columnwidth]{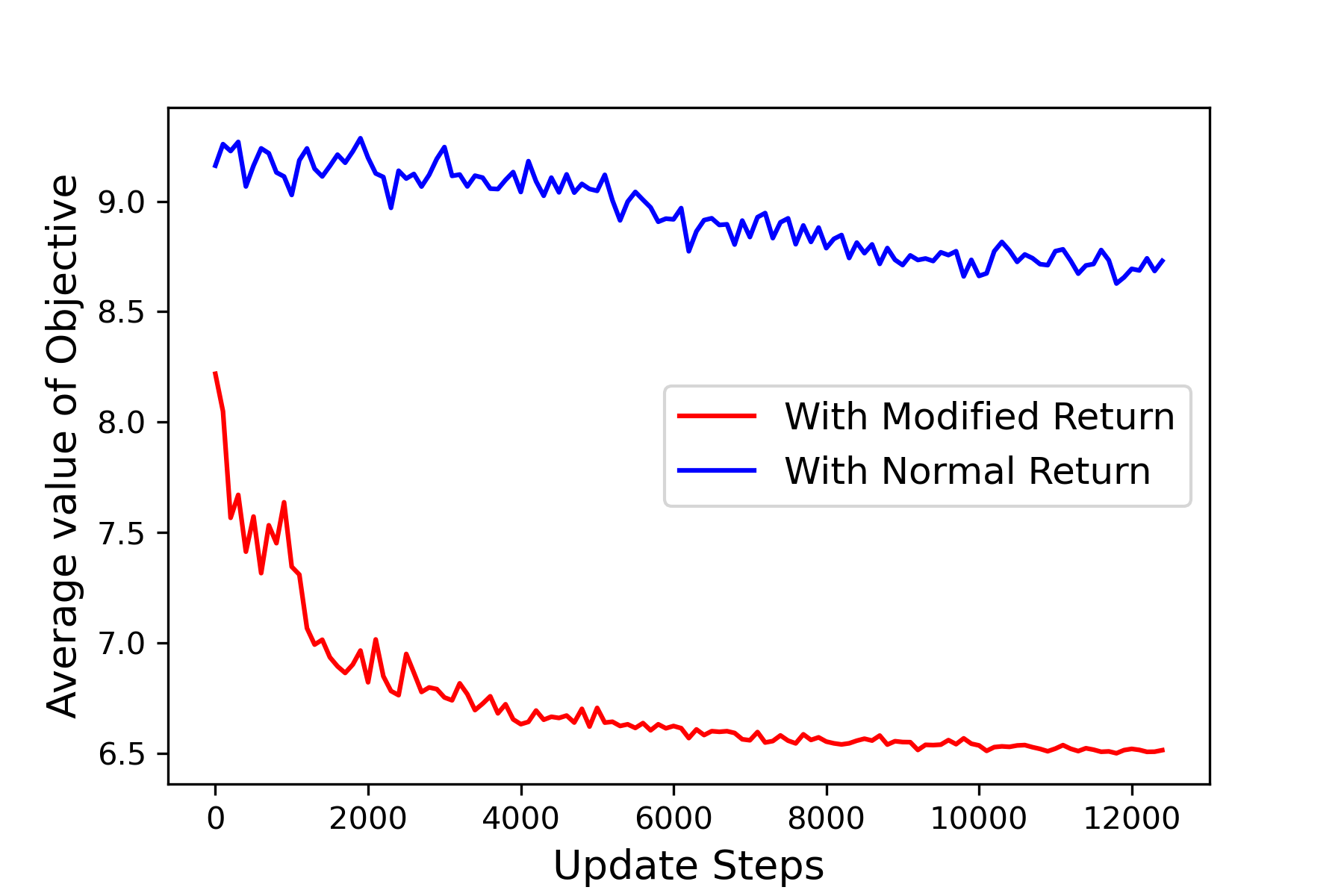}
         \caption{First 12000 training steps of our model with the modified return $G_{t}$ and the ordinary discounted sum return on the dataset of CVRPTW-20. The modified $G_{t}$ shows superior performance with faster convergence speed and lower objective values.}
         \label{fig:gt_vs_return}
     \end{subfigure}
     \hfill
     \begin{subfigure}[b]{0.4\textwidth}
         \centering
         \includegraphics[width=\columnwidth]{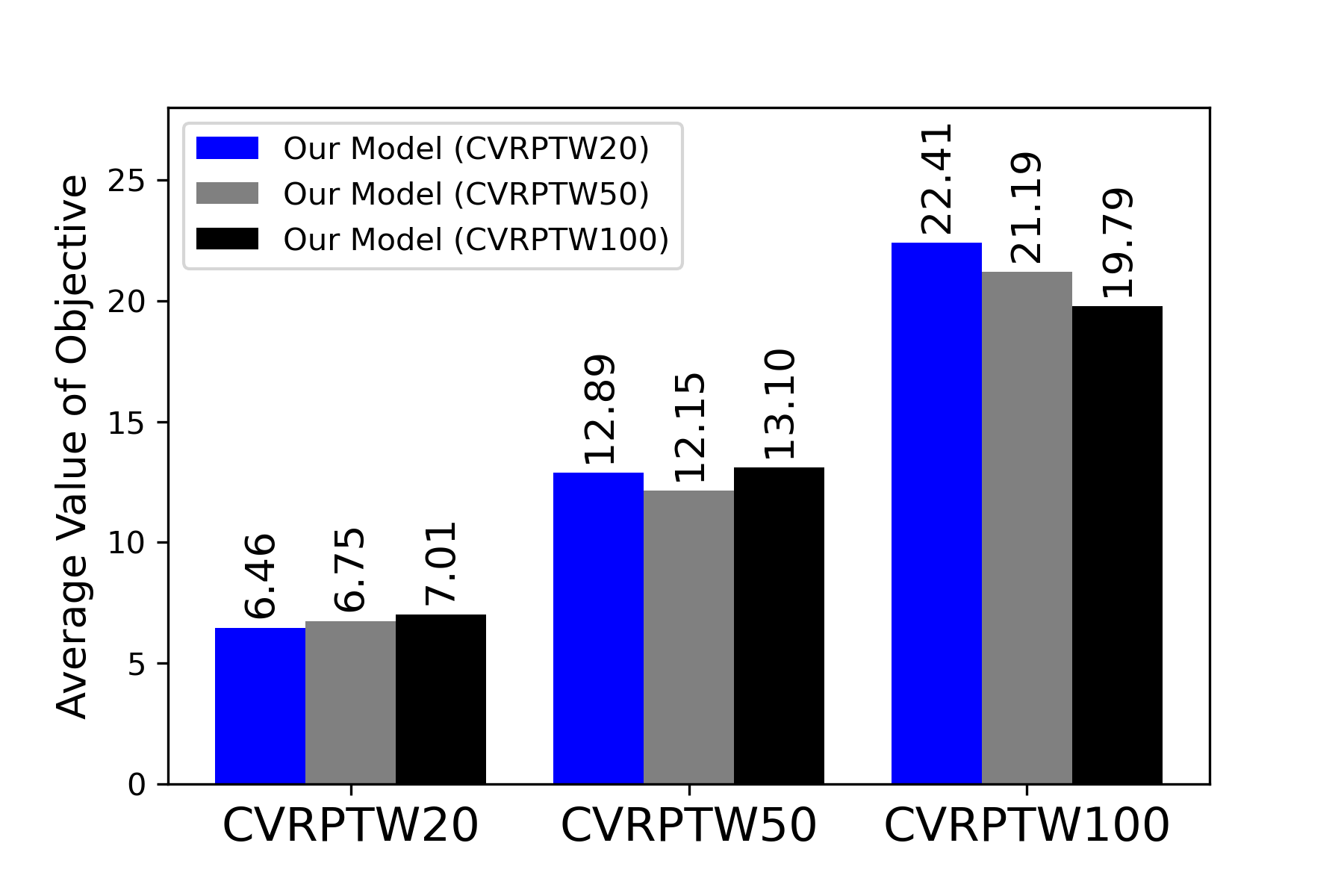}
         \caption{Evaluating the model generalization performance by applying the trained models to solve the same type of problem with different sizes.}
         \label{fig:generalization_cvrptw}
     \end{subfigure}
        \caption{Experiments of the effect of trajectory shaping, $G_{t}$ and a generalization study.}
\end{figure}

Figure \ref{fig:tc_vs_no_tc} shows the effect of trajectory shaping. When trajectory shaping is used, the objective (y-axis) decreases faster and finds smaller values over training epochs (x-axis). This observations indicate the agent should learn easier from manipulated trajectories which are more likely to find better solutions. Figure \ref{fig:gt_vs_return} compares the performance between using the designed $G_{t}$ and the ordinary discounted sum return. The objective oscillates and hardly converges in the ordinary return case. It indicates that the modifications to $G_{t}$ help the agent retrieve more information from trajectories which potentially lead to faster convergence. Table \ref{table:ablation_phi} demonstrates the effect of the initial value of $\lambda$. The reported numbers are evaluated on the same dataset with the CVRPTW(TW) model trained at step 3000. We observe that changing the initial value may affect the balance between the target and cost as well as the convergence speed (solution quality differs at the same number of steps). Empirically, the choice of initial $\lambda$ may be subject to tuning. Figure \ref{fig:generalization_cvrptw} shows the trained model can have decent performance when applying to new test sets with different problem sizes. This potentially promotes practical usage to apply the model trained with one simulated dataset only onto other datasets with different distributions.

\begin{table}[h!]
\centering
\begin{adjustbox}{max width=\columnwidth}
\begin{tabular}{c|ccc}
\hline
 & Obj. & Tgt. & Cost  \\ \hline
(A) Initial $\lambda=1.0$ & 6.59 & 6.56 & 0.033   \\ 
(B) Initial $\lambda=3.0$ & 6.69 & 6.68 & 0.013   \\ 
(C) Initial $\lambda=5.0$ & 6.92 & 6.91 & 0.006   \\ \hline
\end{tabular}
\end{adjustbox}
\caption{The objective, target and cost values for the CVRPTW (TW) model at training step 3000 when $\lambda$ is initialized with different values.}
\label{table:ablation_phi}
\end{table}




\section{Conclusion}
\label{section:conclusion}
In this work, we propose an RL-based framework for soft-constrained VRPs which violating the constraints lead to a finite cost. We incorporate the Lagrangian relaxation technique to move more difficult constraints into the target function and use a constrained policy optimization algorithm to train the RL agent. We apply the framework on VRPs with two types of soft constraints, the time window and the capacity constraint, and also on the VRP with both constraints simultaneously. Comparing to other RL-based work and heuristic solvers, we demonstrate that the proposed method has a competitive performance in finding solutions with a good balance in small travel distance, low constraint violations, and fast inference speed.

Efficiently searching for constraint-satisfied solutions remains to be a challenge in both heuristic and learning-based methods. Though a minor violation of the soft constraints can sometimes be negligible in real-world applications, hard constraints such as pick and delivery are much more difficult to solve. We consider developing an efficient learning-based method to solve hard-constrained problems as our crucial future direction. On the other hand, in many VRP scenarios, MDPs can often be clearly defined. Although the large search space makes solving MDPs difficult, the environment can often be easily defined. Effective utilizing MDP models, such as in model-based RL approaches, might be promising for future research.

\bibliography{ref}
\newpage
\appendix
\section{Other Details of Training}
\label{appendix:train_detail}
\subsubsection{Initial solution strategy}
We use a simple heuristics algorithm \ref{alg:init} to generate initial solutions for CVRP and CVRPTW. The algorithm to generate initial solutions for TSPTW only needs minor adjustments to avoid returning to the depot before finishing visits to customers nodes.

\begin{algorithm}[ht]
\SetKwInOut{Input}{Input}
\SetKwInOut{Hyperparameters}{Parameters}
\SetKwRepeat{Do}{do}{while}
\caption{Generate the initial solution}
\label{alg:init}
\KwIn{the set of depot and customer nodes $\{  n_{0}, n_{1}, n_{2}, \dots, n_{N} \}$, the vehicle capacity $c$, number of customers $N$}
\BlankLine

{Start from the depot and initiate the route $R$ as $[n_{0}]$}\;
{Initiate the set of pending nodes $\mathcal{P}$ as $\{n_{0}, n_{1}, n_{2}, \dots, n_{N}\}$}\;
\Do{$n_{0}$ is not the only node in $\mathcal{P}$}
{
Sort $\mathcal{P}$ by the distance to the previously visited node in ascending order\;
\ForEach{node $n_{i}$ in $\mathcal{P}$}
{
\uIf{$n_{i}$ is the depot node $n_{0}$ and the previous visited node is not the depot}{
Return to the depot, append node $n_{0}$ to $R$, and restore the vehicle capacity to $c$\;
}
\uElseIf{the demand of node $n_{i}$ exceeds the remaining vehicle capacity}{
Return to the depot, append node $n_{0}$ to $R$, and restore the vehicle capacity to $c$\;
}
\Else{
    Go to node $n_{i}$ and remove node $n_{i}$ from $\mathcal{P}$ \;
}
}
}
Return to the depot and append node $n_{0}$ to $R$\;

\textbf{return} $R$
\end{algorithm}

\subsubsection{Hyperparameter settings}
Our model has maximum iteration steps of $200$, $300$, $400$ during training for problem of $n=20, 50, 100$, respectively. To train the model, we use ADAM with a learning rate of 0.0005. $\Phi$ decays from $0.5$ to $0.1$ during training and was fixed to $0.1$ during evaluation. These hyperparameters are set intuitively and we believe doing hyperparameter search should improve the performance. 

\section{Non-linearity of Time Window Constraints}
\label{appendix:nonlinearity}
In this section, we refer to the original \textit{vehicle-flow} formulations used in many exact solution methods such as with Integer Programming (IP) solvers or Branch-and Cut. In these formulations, the decision variable ($x_{ijk}$) is a binary indicator of whether a vehicle $k$ travels from node $i$ to node $j$. An intuition of why time window constraint is non-linear is that $x_{ijk}$ is needed in the calculation of current time, i.e. we need to know the value of both $i$ and $j$ when calculating the travel time between two nodes. Comparing to the capacity constraint, as long as the vehicle arrives at and received the demand of node $j$, we do not need to know which $i$ the vehicle comes from to calculate the current load. In a similar intuition, the constraints in SDVRP and PCVRP are both linear because we can receive the split demand, price or penalty information in these two types of VRPs from the successive node $j$ only. We refer readers to Chapter 5 of \citet{toth} to see the mathematical formulations of VRPTW and the non-linearity in the time window constraint.

\section{Design of the Constrained Cumulative Reward Function}
\label{appendix:design_gt}

Let $s_{t}$ and $s_{t+1}$ represent the state at time $t$ and $t+1$. Without loss of generality, consider the case where there are one constraint only, i.e. $|C|=1$. The the immediate signal when transitioning from $s_{t}$ to $s_{t+1}$ be $(r_{t+1} - r_{t}) + \lambda (c_{t+1} - c{t})$. Then the cumulative signal from $s_{0}$ to $s_{t}$ is 

\begin{equation*}
\begin{split}
    & (r_{1} - r_{0}) + \lambda (c_{1} - c_{0}) + \\
    & (r_{2} - r_{1}) + \lambda (c_{2} - c_{1}) + \ldots + \\
    & (r_{t+1} - r_{t}) + \lambda (c_{t+1} - c_{t}) \\
    & = (r_{t+1} - r_{0}) + \lambda (c_{t+1} - c_{0}).
\end{split}
\end{equation*}


The cancellation works because $\lambda$ is considered fixed as it is updated on a much slower time scale. Therefore, the inside summation in Equation \eqref{eq:cum_constrained_reward} can be roughly seen as the change of value when moving from $s_{0}$ to $s_{t+1}$. The ordinary return function is defined as,

\begin{equation*}
    G_{t} = \sum_{k=0}^{T}\gamma^{k} r_{t+k+1}.
\end{equation*}

Comparing this equation with Equation \eqref{eq:cum_constrained_reward}, we may see that they are similar if replacing $r_{t+k+1}$ with $(r_{t+1} - r_{0}) + \lambda (c_{t+1} - c_{0})$, and replacing summation with a maximum function.


\section{Derivations}
\label{appendix:derivative}

\begin{equation*}
\begin{split}
    \nabla_{\theta} \mathbb{E}_{\pi_{\theta}} \big[ \delta \big] &= \nabla_{\theta} \big[ \mathbb{E}_{\pi_{\theta}} [G_{t}] - \mathbb{E}_{\pi_{\theta}} [V(s)] \big] \\
    &= \nabla_{\theta} \mathbb{E}_{\pi_{\theta}} [G_{t}] \\
    &= \mathbb{E}_{\pi_{\theta}} \big[ \nabla_{\theta} \log P(\pi|\theta) G_{t} \big] \\
    &= \mathbb{E}_{\pi_{\theta}} \bigg[ \sum_{t=0}^{T} \nabla_{\theta} \log \pi_{\theta}(a_{t}|s_{t}) G_{t} \bigg]
\end{split}
\end{equation*}

The first equality comes from linearity of expectation. The second one comes from that $V(s)$ depends on states only and not on policy so it has an expectation of 0. The third and fourth follows from the classic Policy Gradient Theorem.

\begin{equation*}
\begin{gathered}
    \nabla_{\lambda_{c}} \mathbb{E}_{\pi_{\theta}} \big[ G_{t} \big] =  \mathbb{E}_{\pi_{\theta}} \big[ \nabla_{\lambda_{c}} G_{t} \big]
    = - \mathbb{E}_{\pi_{\theta}} \bigg[\gamma^{\tilde{t}-t} \sum_{i=t}^{\tilde{t}} (c_{i} - \epsilon) \bigg], \\
    \tilde{t} = \mathrm{argmax}_{t'} \bigg( \gamma^{t'-t} \Big( \sum_{i=t}^{t'} \big(r_i -  \sum_{C} \lambda_{c} \cdot (c_i - \epsilon) \big) \Big) \bigg). 
\end{gathered}
\end{equation*}

The derivative of $\lambda$ is only taken for a single point $t = \tilde{t}$.
For a specific $\tilde{t}$, i.e. for each piece inside the maximum in $G_{t}$, $G_{t}$ is continuous and differentiable with respect to $\lambda$. The derivative of $G_{t}$ with respect to $\lambda$ is bounded in a neighbourhood of $\tilde{t}$ (a loose bound can be between 0 and a large constant exceeding the sum of distance of all nodes). Following the Dominated Convergence Theorem, the first equality holds given the above conditions.

\section{Sketch of Convergence Conditions}
\label{appendix:convergence}
\subsubsection{Condition 1.} The policy is parameterized using neural networks and we do not learn directly from Q-values.
\subsubsection{Condition 2.} The immediate reward is bounded because it is defined as a difference in the travelling distance or time of two solutions. Given a policy $\pi$, the value of $G_{t}$ is bounded and so does its expected value. Therefore the state value function is bounded for all policy $\pi$ is a reasonable corollary.
\subsubsection{Condition 3.} Each local minimum of the expected value of the cost function indicates a feasible solution. Taking the cost functions defined for capacity and time-window constraint violation as examples, for every routes obtained following policy $\pi$, the global minima for both cost functions are zero, which indicates a feasible solution. 
\subsubsection{Condition 4.} We use constant step sizes $\in (0, 1)$ for updating $\lambda$ and $\theta$ with $\alpha_{\lambda} \ll \alpha_{\theta}$. 

According to \cite{BORKAR2005, tessler2018reward}, the constrained policy optimization algorithm converges to finding fixed-point feasible solutions. 

\end{document}